\documentclass[manuscript, screen]{acmart} %
\usepackage{caption}
\usepackage{subcaption}
\newcommand{\evwindowrank}{I\{ \text{rank}(y|Y_0) \in [\text{rank}(y|Y) \pm T] \}} 
\newcommand{\windowrank}{ \text{rank}(y|Y_0) \in [\text{rank}(y|Y) \pm T]} 
\newcommand{\rankyzero}{\text{rank}(y|Y_0)}
\newcommand{\ranky}{\text{rank}(y|Y)}
\newcommand{\logging}{\pi_0}
\newcommand{\target}{\pi}
\newcommand{\yzeroi}{Y_{0,i}}
\newcommand{\yzero}{Y_{0}}

\newcommand{\EE}{\mathbb{E}}
\newcommand{\PP}{\mathbb{P}}

\newcommand{\interpolestimator}{\hat{\Delta}_{T}(\target| x, \logging, Y_0)}

\newcommand{\ipclick}{\overline{c}(y, \ranky|x)}
\newcommand{\pbclick}{ \text{rel}(y|x) \times \PP(o(y)| \ranky)}

\AtBeginDocument{%
  }

\setcopyright{rightsretained}
\copyrightyear{2022}
\acmYear{2022}
\acmDOI{XXXXXXX.XXXXXXX}

\acmConference[CONSEQUENCES - RecSys 2022]{CONSEQUENCES - RecSys 2022}{September 18--23,
  2022}{Seattle, Washington}
\acmPrice{15.00}
\acmISBN{978-1-4503-XXXX-X/18/06}

\begin{document}

\title[Interpolating the item-position and the position-based model]{Off-policy evaluation for learning-to-rank via interpolating the item-position model and the position-based model}

\author{Alexander Buchholz}
\email{buchhola@amazon.com}
\affiliation{
  \institution{Amazon Music ML}
  \city{Berlin}
  \country{Germany}
}

\author{Ben London}
\email{blondon@amazon.com}
\affiliation{
  \institution{Amazon Music ML}
  \city{Seattle}
  \country{US}
}

\author{Giuseppe Di Benedetto}
\email{bgiusep@amazon.com}
\affiliation{
  \institution{Amazon Music ML}
  \city{Berlin}
  \country{Germany}
}

\author{Thorsten Joachims}
\email{tj@cs.cornell.edu}
\affiliation{
  \institution{Cornell University}
  \city{Ithaca}
  \country{US}
}

\renewcommand{\shortauthors}{Buchholz et al.}

\begin{abstract}
  A critical need for industrial recommender systems is the ability to evaluate recommendation policies offline, before deploying them to production.
  Unfortunately, widely used off-policy evaluation methods either make strong assumptions about how users behave that can lead to excessive bias, or they make fewer assumptions and suffer from large variance.
  We tackle this problem by developing a new estimator that mitigates the problems of the two most popular off-policy estimators for rankings, namely the position-based model and the 
  item-position model.
  In particular, the new estimator, called INTERPOL, addresses the bias of a potentially misspecified position-based model, while providing an adaptable bias-variance trade-off compared to the item-position model.
  We provide theoretical arguments as well as empirical 
  results that highlight the performance of our novel estimation approach. 
\end{abstract}


\maketitle

\vspace{-0.15cm}
\section{Introduction}
Online media streaming platforms rely on highly personalized content recommendation that allows  users to navigate large content pools \cite{mcinerney2018explore, bendada2020carousel}. As the underlying ranking policies constantly evolve, recommendation providers need to experiment offline with new approaches for ranking content before actually deploying and exposing them to the users \cite{gilotte2018offline, jagerman_2019_model_or_intervene}. This serves the purpose of deploying only policies that have a large chance of improving the user experience. 
Deployed ranking policies provide a plethora of interaction logs that can be repurposed to learn and evaluate potentially better policies offline. These logs come in the form of implicit feedback, i.e., records of past interaction behavior, linked to information about the user, the context and the items to recommend. 
Off-policy evaluation of new policies on historic data requires adequate strategies to deal with biases coming from (i) the nature of user interaction and (ii) the logging policy. A prominent example of these biases is %
position bias \cite{joachims2017accurately} (content that is not ranked in the most visible positions is less likely to be seen). 
We focus on two popular classes of estimators that take different approaches to correcting for presentation bias.
The first class, in the case of full visibility of all items, does not rely on explicit randomization, but models the randomness in user behavior. The most common model is the position-based model (PBM), which assumes that observed clicks on content factorize into relevance (depending on the item only) and the visibility of the content (depending on the position only). The PBM has been successfully used in practice and various methods have been developed for estimating position bias curves \cite{agarwal2019estimating, ai2018unbiased, joachims2017unbiased, wang2018position, ruffini2022modeling}. In realistic scenarios, the position bias curve (i.e., examination probabilities) must be estimated and so will almost always be approximate, which can lead to a bias in the evaluation (even if true user behavior factors as assumed by the PBM). 
The other class of estimators requires explicit randomization during data collection. The most popular estimator in this class is the item-position model (IPM) \cite{li2018offline}. Unlike the PBM, the IPM allows clicks to be a function of interactions between the recommended content and its display position and does not require the estimation of a position bias curve. 
Since it uses explicit randomization, its bias is typically lower but at the expense of increased estimator variance.

To obtain a better balance of bias and variance under realistic conditions, we propose a new estimator that interpolates between the PBM and the IPM in the full visibility setting. 
We show that this estimator is always unbiased for a correctly specified PBM, but can have better variance than both the PBM and the IPM. More importantly, for a misspecified PBM, 
our new estimator is based on the idea that the position-based model might provide a good approximation to local behavior, i.e., small differences in ranking position are properly modeled, but large jumps from the top to the bottom of a list lead to unreliable examination probabilities. 
The IPM calibrates the PBM by computing probabilities that an item is within the range (i.e., window size) of another one, if the PBM is correctly specified. The window size serves as a tuning parameter that allows to trade off potentially high variance IPM estimation with a potentially more biased PBM. We show empirically that this leads to reduced error and hence provides a more precise estimation strategy.

\vspace{-0.15cm}
\section{Related work} \label{sec:related_work}
Our work provides a solution to off-policy evaluation of ranking models using implicit feedback \cite{swaminathan2017off, wang2016learning, joachims2017unbiasedltr}, that rely on some form of click model \cite{chuklin2015click} to achieve unbiased offline evaluation. The arising bias comes from user behavior such as 
position bias, trust bias, and selection bias, see for instance \cite{joachims2017accurately, joachims2007evaluating}. 
A popular click model is the position-based model, that needs an estimated position bias curve \citep{wang2016learning}. 
We make extensive use of Li et al.'s \cite{li2018offline} survey of click models for offline evaluation.
Other approaches go beyond the full visibility setting such as policy aware approaches \cite{oosterhuis2020policyaware}.%
\vspace{-0.15cm}
\section{Background} \label{sec:background}
We are interested in estimating the total number of clicks %
for a target policy $\target$ using recorded interactions from a production policy $\logging$. 
The quantity of interest is 
$ \Delta_\target = \EE_x \EE_c \EE_{Y \sim \target(\cdot|x)}\left [ \sum_{y \in Y} c(y) \right ], $
where we compute expectations over context features $x$, received clicks $c$ and exposed rankings $Y$ that contain items $y$. 
We make use of logging data of the form $\mathcal{D} =  \left\{x_i, \yzeroi, c(\cdot | \yzeroi ), \logging \right\}_{i=1}^n$ for $n$ different queries, where $\yzeroi \sim \logging(\cdot|x_i)$ and $c(y | \yzeroi ) \in \{0,1\}$ indicates which items $y \in \yzeroi$ received a click. 
Our suggested estimator has the form 
\begin{eqnarray} \label{eq:estimator}
  \hat{\Delta}(\target| \mathcal{D}) = \frac{1}{n} \sum_{i=1}^n \sum_{y \in \yzeroi} w \times c(y| \yzeroi). 
\end{eqnarray} 
The inverse propensity score (IPS) weight $w$ corrects the fact that logging and policy are different and accounts for position biases. We will specify the form of $w$ depending on the underlying assumptions. 
The expectation of a click on item $y$ is 
\begin{eqnarray}
  \EE_c [c(y|Y)|x] = 
  \begin{cases}
    \ipclick, & \text{for the item-position model,}\\
    \pbclick, & \text{for the position-based model.}
  \end{cases}
\end{eqnarray}
Under the IP model a click depends on item $y$ and its position, defined by its rank, whereas the PB model assumes a click factorizes into relevance $\text{rel}(y)$ and observation probability $\PP(o(y) | \ranky)$ of the item, where $o(y)$ indicates if an item was examined. 

\textbf{Position-based model.}
The position-based model corrects the fact that 
not all positions have equal probability of being observed by the user.
By weighting clicks using a position bias curve the examination behavior of the user is taken into account.
A position bias curve $p$ quantifies the probability of an item being observed in a given position, i.e., $p_k = \mathbb{P}(o(y)| \ranky = k),$ where $o(y)$ denotes the event that item $y$ is observed in the position it was displayed in. 
The corresponding IPS weight (in the full-visibility setting) is
\begin{eqnarray}
  w_{pbm}(y| Y, Y_0) = \frac{p_{\ranky}}{p_{\rankyzero}}.
\end{eqnarray}
Thus, clicks are weighted according to the visibility ratio of items under logging policy and target policy. 

\textbf{Item-position model.}
The item-position model \cite{li2018offline} does not require a position bias curve. It uses directly the propensities of the logging policy, denoted $\PP(\rankyzero = \cdot | x, \logging)$. The propensities quantify the probability with which an item is displayed in a given rank. The resulting IPS weight is
\begin{eqnarray} \label{eq:ip}
  w_{ip}(y| Y, Y_0) = \frac{I\{ \rankyzero = \ranky \}}{\PP(\rankyzero = \ranky | x, \logging)}.
\end{eqnarray}
Here, we assume that the 
target policy is deterministic, and the logging policy is stochastic. The IP model assigns a weight of $0$ if target and logging position mismatch and weights up rewards where target and logging rank agree. 

\section{INTERPOL estimator} \label{sec:contribution}
The position-based model captures overall user behavior, but it can be biased. The item-position model makes fewer modeling assumptions that may lead to bias, but it can have high variance. 
We suggest an interpolation between the two approaches. 
We check if the position where the logging policy ranks an item 
$\rankyzero$ falls inside a window of size $T$
around the position $\ranky$ where the target policy ranked it.
The size of the window is controlled by the interpolation parameter $T$, and we denote 
the event of y being inside the window by $\evwindowrank$. 
Our novel estimator is based on the weight
\begin{eqnarray} \label{eq:interpolation}
  w_{T}(y|Y, Y_0) =  \frac{\evwindowrank}{P( \windowrank | x, \logging)} \times \frac{p_{\ranky}}{ p_{\rankyzero}}.
\end{eqnarray} 
If $T = 0$, then the position of an item has to be identical under the target and logging policy in order to provide a non-zero weight for observed rewards. In this case we recover the IPM as the PBM part is either 1 or ignored in case of a miss-match. For $T > 0$ position bias weights are limited to at most $T$ positions apart. If $T$ is equal to the length of the displayed list $|Y|$ in the full visibility setting, the resulting denominator is equal to $1$ as $P( \windowrank | x, \logging) = 1$ and the PBM is recovered. 
We denote by $\interpolestimator$ the estimator that uses the weights as defined in \eqref{eq:interpolation} inside the generic estimator in \eqref{eq:estimator}. 

\vspace{-0.15cm}
\begin{proposition} \label{prop:unbiasedness}
  $\interpolestimator$ is an unbiased estimator of $\Delta_\target$ for all window sizes $T$ if the logging data is generated from a known logging policy $\logging$ with full support, under the position-based model with a known position bias curve $p$. 
  (See the appendix for a proof.)
\end{proposition} 

\vspace{-0.3cm}
\section{Experiments} \label{sec:experiments}
We illustrate our off-policy evaluation approach INTERPOL with experiments on a synthetic data set that is simulated in a controlled environment using $5,000$ data points. We highlight the impact of varying (i) the position bias misspecification (using powers of the true curve); (ii) the different window sizes and (iii) the randomization of the logging policy (via random position swaps, which are controlled by a parameter called stay probability) on our offline evaluation. Implementation details are available in the appendix. 
\begin{figure}[t!] 
  \centering
  \begin{subfigure}[b]{0.35\textwidth}
    \centering
    \includegraphics[width=\textwidth]{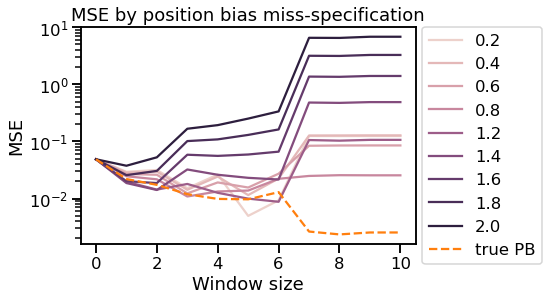}
    \caption{Stay probability set to $0.95$. }
    \label{fig:mse_varying_pb_n_selected_10}
  \end{subfigure}
  \hfill
  \begin{subfigure}[b]{0.3\textwidth}
  \centering
  \includegraphics[width=\textwidth]{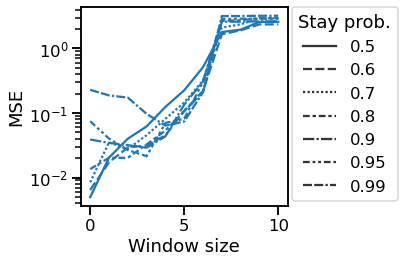}
  \caption{Misspecified pb curve set to $p^{1.8}$.}
  \label{fig:varying_stay_prob_n_selected_10}
  \end{subfigure}
  \begin{subfigure}[b]{0.3\textwidth}
    \centering
    \includegraphics[width=\textwidth]{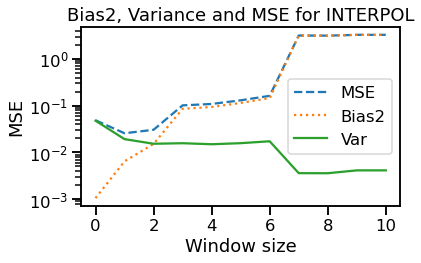}
    \caption{Bias-variance decomposition for pb curve $=p^{1.8}$, stay probability $=0.95$.}
    \label{fig:mse_var_bias_n_selected_10}
    \end{subfigure}
    \vspace{-0.15cm}
  \caption{Estimated MSE for different position bias curves in Figure \ref{fig:mse_varying_pb_n_selected_10}. Figure \ref{fig:varying_stay_prob_n_selected_10} highlights different randomization strengths of the logging policy with the resulting MSE over different window sizes. Figure \ref{fig:mse_var_bias_n_selected_10} illustrates a bias-variance decomposition of the MSE for the position bias curve set to $p^{1.8}$ and the stay probability of the logging policy randomization set to $0.95$.}
  \vspace{-0.5cm}
\end{figure}

\textbf{Results.}
When it comes to the interpolation between the IPM (window size 0) and the PBM (window size 10) we see that the arising bias-variance trade-off in Figure \ref{fig:mse_varying_pb_n_selected_10} is impacted by the misspecification of the position bias curve. As the window size goes up (left to right) the MSE of the estimator first decreases due to a reduction in variance and then increase due to an increase in bias. For all levels of misspecification we eventually end at a clearly biased PBM estimator. For small powers (below $1$) the misspecification of the PBM actual acts as weight clipping, which seems to be beneficial in terms of MSE for small window sizes. There seems to be a region around window size 1 to 6, where the MSE is lowest. Consequently, the estimator that offers the best bias-variance trade-off is INTERPOL with a properly chosen window size. Interestingly, the interpolation of INTERPOL can also reduce variance even if the correct position bias curve is used, as highlights Figure \ref{fig:mse_varying_pb_n_selected_10}, where a large window size of 8 leads to favorable MSE. 
When varying the randomization of the logging policy, see Figure \ref{fig:varying_stay_prob_n_selected_10}, 
we also identify a favorable bias-variance trade-off for a window size between $2$ and $6$ for a weak randomization of the logging policy (stay probability above $0.9$). For stronger randomization the IPM has a lower MSE. 
Figure \ref{fig:mse_var_bias_n_selected_10} illustrates a bias-variance decomposition of the estimator. For a window size larger than $2$ the bias starts dominating the MSE.

\section{Discussion and conclusion} \label{sec:conclusion}
We have introduced a novel off-policy estimator, called INTERPOL, for learning-to-rank in the full visibility setting that interpolates the IPM and the PBM. INTERPOL has a favorable MSE, even when the PBM is correctly specified. 
In future work, we plan to include the top-k setting, investigate the misspecified version of the PBM %
and study the MSE of different window size $T$ from a theoretical perspective. We also want to extend experiments and provide methods for choosing the best window size, using ideas based on \cite{su2020adaptive}. 
Finally, we aim to develop off-policy learning approaches based on our interpolation idea.

\bibliographystyle{ACM-Reference-Format}
\bibliography{literature}

\newpage

\appendix
\section{Theoretical results}
In this section we provide the proof of our main result.

\subsection{Unbiasedness of the interpolating estimator under the position-based model}
We show that the interpolating estimator is unbiased if user interactions are actually coming from a position-based model, and we dispose of the correct position bias curve under the full visibility setting. 
\paragraph{Proof of Proposition \ref{prop:unbiasedness}}
\begin{proof}
  We focus on showing that $\interpolestimator$ is unbiased for a single sample $x$. The generalization using the distribution over contexts $x$ is straightforward. 
  We evaluate
  \begin{eqnarray}
   && \EE \left[ \interpolestimator \right]  \\
     &=& \EE \left[
    \sum_{y \in \yzero} \frac{\evwindowrank}{P( \windowrank | x, \logging)} \times \frac{\sum_{k=1}^K I\{ \ranky = k \} \times p_k}{\sum_{k=1}^K I\{ \rankyzero = k \} \times p_k} \times c(y| \yzero) \right], \label{eq:line_proof_unbiasedness_1}\\
    &=& 
    \EE_c \EE_{\logging} \left[
    \sum_{y \in \yzero} \frac{\evwindowrank}{ \PP ( \windowrank | x, \logging)} \times \frac{p_{\ranky}}{p_{\rankyzero}} \times c(y| \yzero) \right], \label{eq:line_proof_unbiasedness_2} \\
    &=& 
    \EE_{\logging} \left[
    \sum_{y \in \yzero} \frac{\evwindowrank}{ \PP ( \windowrank | x, \logging)} \times \frac{p_{\ranky}}{p_{\rankyzero}} \times \EE_c  c(y| \yzero) \right],  \label{eq:line_proof_unbiasedness_3} \\
    &=& 
    \EE_{\logging} \left[
    \sum_{y \in \yzero} \frac{\evwindowrank}{ \PP ( \windowrank | x, \logging)} \times \frac{p_{\ranky}}{p_{\rankyzero}} \times  \text{rel}(y) \PP(o(y)| \rankyzero) \right], \label{eq:line_proof_unbiasedness_4} \\
    &=& 
    \EE_{\logging} \left[
    \sum_{y \in \yzero} \frac{\evwindowrank}{ \PP ( \windowrank | x, \logging)} \times p_{\ranky} \times \text{rel}(y) \right], \label{eq:line_proof_unbiasedness_5} \\
    &=& 
    \sum_{y \in \yzero} \frac{ \EE_{\logging} \left[ \evwindowrank \right]}{ \PP ( \windowrank | x, \logging)} \times p_{\ranky} \times  \text{rel}(y),  \label{eq:line_proof_unbiasedness_6} \\
    &=& \sum_{y \in \yzero} \frac{ \PP ( \windowrank | x, \logging)}{ \PP ( \windowrank | x, \logging)} \times p_{\ranky} \times \text{rel}(y),  \label{eq:line_proof_unbiasedness_7} \\
    &=& \sum_{y \in \yzero} p_{\ranky} \times \text{rel}(y) = \EE_c \EE_{\target} \left[ \sum_{y \in Y} c(y) \right]. \label{eq:line_proof_unbiasedness_8}
  \end{eqnarray}
  Line \eqref{eq:line_proof_unbiasedness_1} is obtained by using the definition of \eqref{eq:estimator}. Line \eqref{eq:line_proof_unbiasedness_2} uses the definition of the position bias weights and decomposes the expectation. Line \eqref{eq:line_proof_unbiasedness_3} pulls the expectation of the click model inside the sum, exploiting the linearity of the expectation and the fact that clicks do not interact by assumption of the PBM. Line \eqref{eq:line_proof_unbiasedness_4} uses the definition of a click under the PBM and in line \eqref{eq:line_proof_unbiasedness_5} we simplify the production of examination probability. In Line \eqref{eq:line_proof_unbiasedness_6} we pull the expectation with respect to the logging policy inside the sum and line \eqref{eq:line_proof_unbiasedness_7} evaluates this expectation. Finally line \eqref{eq:line_proof_unbiasedness_8} simplifies the expression and uses the previous identities in reverse using the position-based model under the target policy $\target$. 
\end{proof}

\section{Experiments}
We describe our experiment in more detail and provide more results on the estimator in the full visibility setting.

\begin{figure}[ht!] 
    \includegraphics[width=0.5\textwidth]{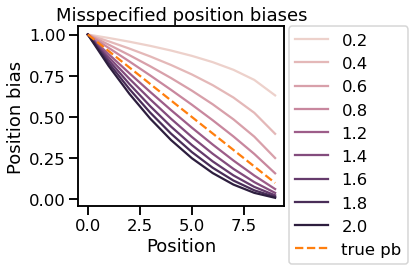}
    \caption{Position bias curves used in the experiments.}
    \label{fig:pbcurves}
\end{figure}

\textbf{Data generation for the toy experiment}
Our synthetic data generation uses a toy model that allows to easily compute expectations of the true reward as well as to control the strength of logging policy randomization and the level of position bias misspecification. This experimental set-up is not supposed to be realistic, but rather to study properties of INTERPOL easily. 
We generate $5,000$ observations from a synthetic logging policy that ranks $K=10$ different actions. 
The true position bias curve is given as $p = [1, 0.9, 0.8, \cdots, 0.1]$, the biased curve is defined as $p^{x}$ component wise, where $x \in [0.2, 0.6, 0.8, 1, 1.2, 1.4, 1.6, 1.8, 2.0]$. We illustrate the curves for $p^x$ in Figure \ref{fig:pbcurves}. 
The relevant items are items $[1, 2, 4, 7]$ and the logging policy $\logging$ orders items $[6, 0, 3, 1, 4]$ at the top and items $[7, 5, 2]$
at the bottom of the list. The other items are displayed in arbitrary order. Additionally, the logging policy swaps the ranked items randomly, where every item has a probability of $q\%$ of staying in its original position and a probability of $(100-q\%) / 9$ of being ranked in all other positions. We set these probabilities to $[50\%, 55\%, 60\%, 70\%, 80\%,  90\%, 95\%, 99\%]$. We denote this probability \textit{stay probability}. 
The target policy $\target$ deterministically ranks items 
$[7, 0, 3, 1]$ at the top and items $[2, 4]$ at the bottom. 
A positive reward of $1$ is generated for the relevant items and this reward is revealed according to the examination probability (i.e., position bias curve $p$). 
Hence, the expected reward for the target policy (under full visibility) is 
$1\times p_0+1\times p_3 + 1\times p_8 + 1 \times p_9 = 2$, where $p_0 = 1, p_3 = 0.7, p_8=0.2, p_9=0.1$. %

\begin{figure}
    \centering
    \includegraphics[width=0.8\textwidth]{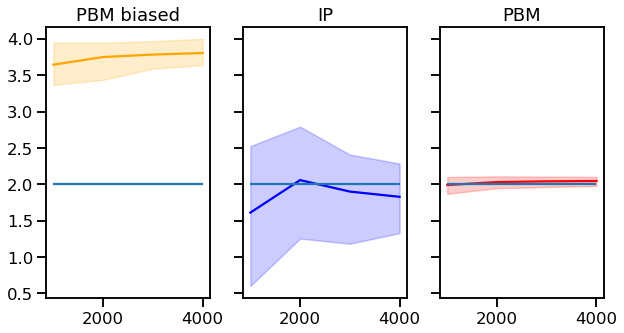}
  \caption{Estimated reward for different estimators over different data set size in the full visibility setting. Stay probability set to $95\%$ and the misspecified position bias curve is $p^{1.8}$.}
  \label{fig:ip_pb_toy}
\end{figure}

\begin{figure}
  \centering
  \includegraphics[width=1\linewidth]{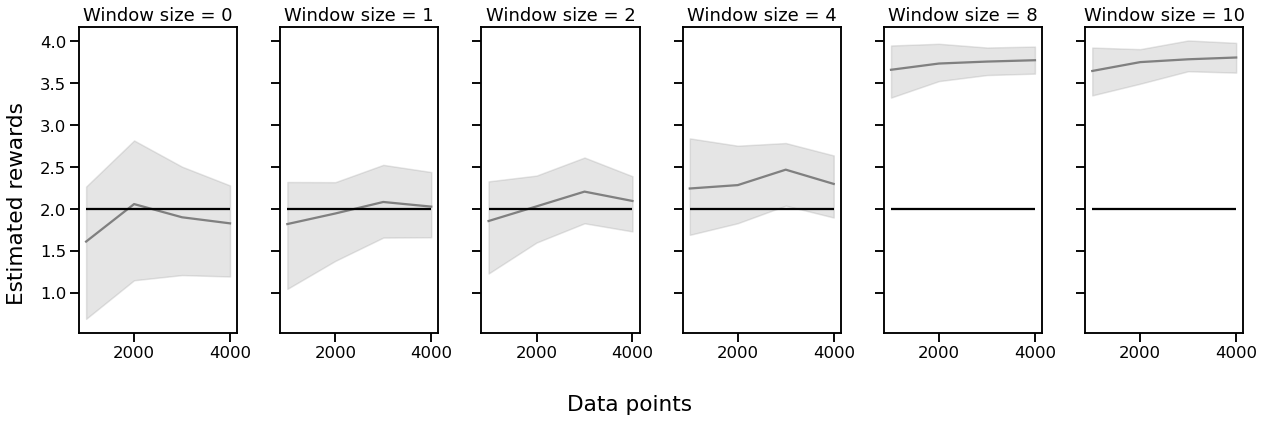}
  \caption{Interpolation of INTERPOL in the full visibility setting. Stay probability set to $95\%$ and the misspecified position bias curve is $p^{1.8}$.}
  \label{fig:convergence_estimators_n_selected_10_windows}
\end{figure}

Figure \ref{fig:ip_pb_toy} highlights the behavior of the item-position model and the position-based model. The left-hand figure uses a misspecified position bias curve and clearly the resulting estimate is severely biased. The IP model (middle figure) exhibits higher variance but the true reward (straight line) lies inside the 95 \% confidence interval of the IPM estimator. As the data set size increase, the estimator gets more precise and the confidence intervals shrink around the true value ($2.0$). The PBM (right-hand figure) that uses the correct position bias curve estimates the reward correctly and has less variability than the IP model. 
For illustrative purpose we also show the interpolation over different window sizes in Figure \ref{fig:convergence_estimators_n_selected_10_windows}.

\end{document}